\documentclass[letterpaper, 10 pt, conference]{ieeeconf}  %

\IEEEoverridecommandlockouts                              %

\makeatletter
\let\NAT@parse\undefined
\makeatother

\usepackage{amsmath,amsfonts}
\usepackage{algorithmic}
\usepackage{algorithm}
\usepackage{array}
\usepackage[caption=false,font=normalsize,labelfont=sf,textfont=sf]{subfig}
\usepackage{textcomp}
\usepackage{stfloats}
\usepackage{url}
\usepackage{verbatim}
\usepackage{graphicx}
\usepackage{cite}
\usepackage{xcolor}
\usepackage{booktabs}
\usepackage{hyperref}
\usepackage{afterpage}
\usepackage{caption}
\usepackage{overpic}
\usepackage{mathtools}

\newcommand{\Tau}{\mathrm{T}}

\title{\LARGE \bf
Scalable Real2Sim: Physics-Aware Asset Generation \\ Via Robotic Pick-and-Place Setups
}

\author{Nicholas Pfaff$^{1}$, Evelyn Fu$^{1}$, Jeremy Binagia$^{2}$, Phillip Isola$^{1}$, and Russ Tedrake$^{1}$%
}

\begin{document}

\twocolumn[{
\renewcommand\twocolumn[1][]{#1}
\maketitle

\includegraphics[width=\linewidth, keepaspectratio]{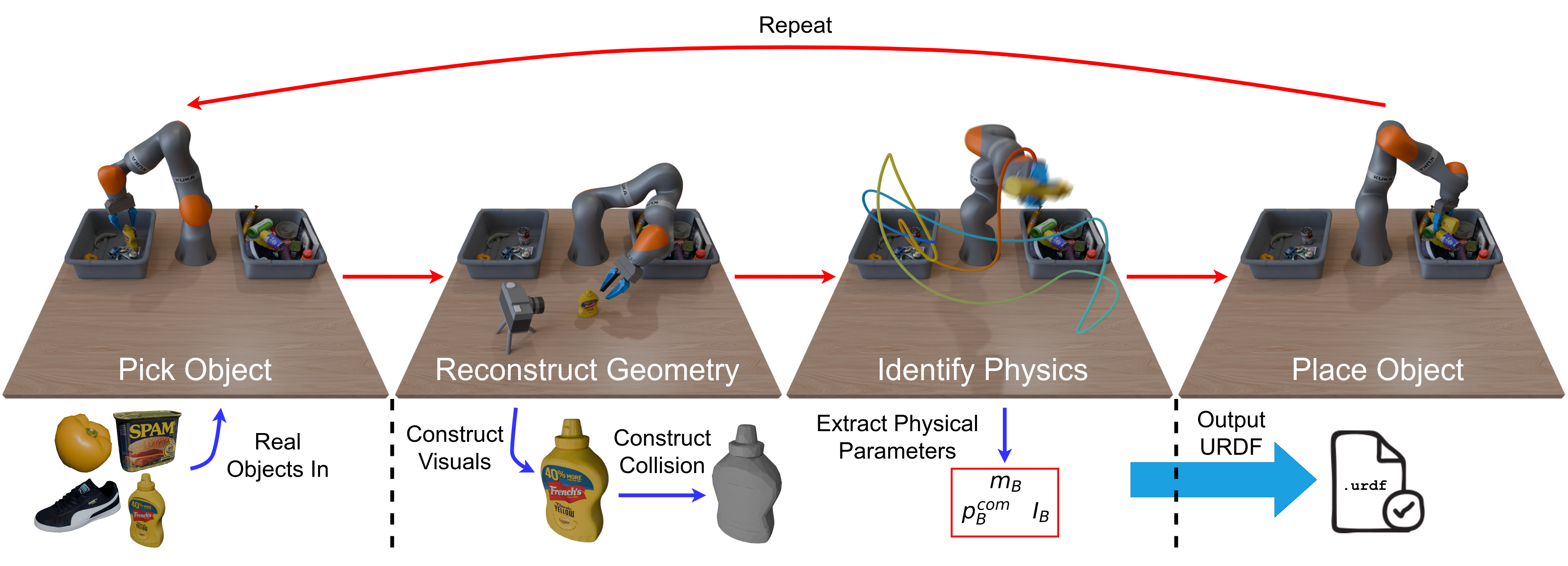}
\captionof{figure}{\textbf{An overview of our system.} Objects are placed in the first bin, where the robot picks them up and reconstructs their geometries by moving them in front of a static camera while re-grasping to reduce occlusions (Section \ref{visual_geometry} \& \ref{collision_geometry}). Next, the robot identifies the object's physical parameters by following a trajectory designed to be informative for the inertial parameters (Section \ref{system_id}). Finally, it places the object into the second bin and repeats the process with the next object. The extracted geometric and physical parameters are combined to generate a complete, simulatable object description.}
\vspace{1em}
\label{system_diagram}
}]

\thispagestyle{empty}
\pagestyle{empty}

\footnotetext[1]{Nicholas Pfaff, Evelyn Fu, Phillip Isola, and Russ Tedrake are with the Massachusetts Institute of Technology, Cambridge, MA, USA, {\tt\small \{nepfaff, evelynfu, phillipi, russt\}@mit.edu}}%
\footnotetext[2]{Jeremy Binagia is with Amazon Robotics, \newline {\tt\small jbinagia@amazon.com}}
\renewcommand{\thefootnote}{\arabic{footnote}}
\setcounter{footnote}{0}

\begin{abstract}

Simulating object dynamics from real-world perception shows great promise for digital twins and robotic manipulation but often demands labor-intensive measurements and expertise. We present a fully automated Real2Sim pipeline that generates simulation-ready assets for real-world objects through robotic interaction. Using only a robot’s joint torque sensors and an external camera, the pipeline identifies visual geometry, collision geometry, and physical properties such as inertial parameters. Our approach introduces a general method for extracting high-quality, object-centric meshes from photometric reconstruction techniques (e.g., NeRF, Gaussian Splatting) by employing alpha-transparent training while explicitly distinguishing foreground occlusions from background subtraction. We validate the full pipeline through extensive experiments, demonstrating its effectiveness across diverse objects. By eliminating the need for manual intervention or environment modifications, our pipeline can be integrated directly into existing pick-and-place setups, enabling scalable and efficient dataset creation.
Project page (with code and data): \url{https://scalable-real2sim.github.io/}.

\end{abstract}

\section{Introduction}

Physics simulation has been a driving force behind recent advances in robotics, enabling learning in simulation before deploying in the real world \cite{sim2real_survey_2020}. This paradigm, termed \emph{Sim2Real}, shows strong potential for machine learning approaches like reinforcement learning, which require large interaction datasets that are easier to collect in simulation than in reality. This approach will become increasingly valuable as robotics shifts toward foundation models, which require even larger training datasets \cite{foundation_models_in_robotics}. However, Sim2Real typically requires manually replicating real-world scenes by curating object geometries and tuning dynamic parameters like mass and inertia. This process is time-consuming, requires expertise, and is challenging to scale.

We propose an automated pipeline to generate dynamically accurate simulation assets for real-world objects. Inspired by prior work in \emph{Real2Sim} \cite{memmel2024asid, real2sim2real_hao_su, google_scanned_objects, pulkit_real2sim}, our approach reconstructs an object’s geometry and identifies its physical properties to create a complete simulation asset. Unlike prior methods, which primarily focus on either dynamic parameter identification \cite{memmel2024asid} or geometric reconstruction \cite{real2sim2real_hao_su, pulkit_real2sim}, our pipeline combines both, producing assets with visual and collision geometry as well as physical properties like mass and inertia.
One could view this as a kind of \emph{physics scanner}; analogous to how 2D scanners digitize documents and 3D scanners reconstruct geometric shapes, our method scans objects to create physically realistic digital twins.
Furthermore, our method autonomously interacts with real-world objects to generate these assets and is designed to work with existing pick-and-place setups without extra hardware or human intervention. Future versions could allow robots to passively generate object assets as a byproduct of tasks like warehouse automation, where small adjustments to planned trajectories could provide data for the same model across multiple picks.

Our pipeline integrates seamlessly into a standard pick-and-place setup; we give an example with two bins and a robotic arm. The robot picks an object from the first bin, interacts with it to gather data for asset creation, and places it in the second bin. This process runs autonomously until the first bin is empty.
The robot begins by moving the object in front of a static RGBD camera, re-grasping as needed to expose all sides and ensure the entire surface is captured.
Background and gripper pixels are masked using video segmentation, while object tracking estimates object poses in the camera frame. These poses, along with the image data, serve as inputs to a 3D reconstruction method.
We introduce a general recipe for obtaining object-centric triangle meshes from arbitrary photometric reconstruction techniques such as NeRF \cite{nerf} and Gaussian Splatting \cite{gaussian_splatting}. Our training procedure supports reconstruction from input views where the object moves and may be partially occluded, such as by a gripper.
The collision geometry is then derived via convex decomposition, simplifying and convexifying the shape for simulation purposes. Lastly, the robot follows an excitation trajectory designed to maximize information gain, collecting joint position and torque data to identify the object's physical properties via convex optimization.
To handle arbitrary constraints in trajectory design, we introduce a custom augmented Lagrangian solver that solves its subproblems using black-box optimization. Unlike gradient-based solvers, which often struggle with local minima when additional constraints are introduced due to the numerical challenges of the information objective, our approach better handles complex constraints such as collision avoidance.

We evaluate our pipeline and its components, demonstrating millimeter-level reconstruction accuracy and the ability to estimate mass and center of mass with only a few percent error.
We use our pipeline to generate an initial small dataset, showcasing its capability to create complete simulation assets autonomously. Our results highlight its potential as a scalable solution for future large-scale dataset collection.

In summary, our key contributions are:
\begin{itemize}
    \item A fully automated pipeline that generates complete simulation assets (visual geometry, collision geometry, and physical properties) using pick-and-place setups without hardware modifications or human intervention.
    \item A general recipe for obtaining object-centric triangle meshes from photometric reconstruction methods such as NeRF for moving, partially occluded objects by employing alpha-transparent training and distinguishing foreground occlusions from background subtraction.
    \item Practical implementations of optimal experiment design and physical parameter identification, allowing arbitrary robot specifications and leveraging a custom augmented Lagrangian solver for finding excitation trajectories under arbitrary constraints.
    \item Extensive real-world experiments validating the effectiveness of the pipeline and its individual components.
    \item A benchmark dataset of 20 assets generated by our pipeline, including raw sensor observations used in their creation. This dataset enables researchers to improve aspects of our pipeline, such as object tracking, geometry reconstruction, and inertia estimation, without requiring access to a robotic system.
\end{itemize}

\section{Related Work}

\textbf{Real2Sim in Robotics.}
Real2Sim \cite{memmel2024asid, real2sim2real_hao_su, google_scanned_objects, pulkit_real2sim} involves generating digital replicas of real-world scenes to reduce the Sim2Real gap. Wang et al. \cite{real2sim2real_hao_su} construct object meshes from depth images, providing a foundation for digital reconstruction, though their approach primarily focuses on static objects. Downs et al. \cite{google_scanned_objects} achieve high-accuracy geometry scans using custom hardware, offering precise shape representations but requiring manual object handling and omitting physical properties. Torne et al. \cite{pulkit_real2sim} present a user interface for creating digital twins, streamlining the modeling process while relying on human input for asset creation.
Our approach extends previous work by capturing physical properties in addition to geometric and visual ones, producing complete simulation-ready assets without manual intervention.

\textbf{Generative AI for Asset Generation.}
An alternative to Real2Sim for asset creation is the use of generative AI \cite{katara2023gen2sim, liu2023zero1to3, LumaGenie}. Gen2Sim \cite{katara2023gen2sim} employs diffusion models to synthesize 3D meshes from a single image and queries large language models for object dimensions and plausible physical properties. While generative AI methods are easier to scale for large-scale asset creation, the Real2Sim approach is preferable when assets are intended to accurately reflect real-world objects rather than being AI-imagined approximations. The observations from \cite{wei2025understandingcotraining} suggest that accurate physical parameters are important for simulation-trained policies to transfer to the real world for nonprehensile manipulation.

\textbf{3D Reconstruction.}
3D reconstruction methods focus on reconstructing geometry from real-world observations such as RGB and depth images. Modern methods produce implicit representations \cite{nerf, gaussian_splatting}, which excel at high-quality rendering but do not always yield structured geometry \cite{neuralangelo}. We propose a general recipe to obtain object-centric triangle meshes from arbitrary photometric reconstruction methods by employing alpha-transparent training \cite{instantngp} while explicitly distinguishing foreground occlusions from background subtraction.

\textbf{Inertial Parameter Estimation \& Experiment Design.}
The identification of dynamic parameters for robotic systems and payloads is well-studied; see \cite{robot_id_survey_2024} for a recent review. There are two main approaches to payload identification: (1) identifying the robot's parameters using joint-torque sensors, then re-identifying them with the payload to compute payload parameters as the difference \cite{payload_id_of_industrial_manipulators}, and (2) using a force-torque sensor at the end-effector to directly identify payload parameters \cite{fast_object_inertial_id}. We focus on the first approach as it avoids the need for a wrist-mounted sensor and provides robot parameters for precise control. However, the second could also be used in setups without joint-torque sensors.

A key challenge in inertial parameter estimation is ensuring physical feasibility in situations where we have limited data or some parameters are unidentifiable \cite{physical_consistent_LMI, geometric_robot_dynamic_id}. \cite{physical_consistent_LMI} formulates linear matrix inequality (LMI) constraints for physical consistency, and \cite{geometric_robot_dynamic_id} adds additional regularization techniques. We implement these methods in Drake \cite{drake}, enabling identification for any robot specified by URDF, SDFormat, or MJCF description files.

Another challenge is collecting informative data. Optimal excitation trajectory design \cite{birdy, leeOptimalExcitationTrajectories2021, Tian2024ExcitationTO} tackles this by generating trajectories that maximize parameter information gain. Prior works often hard-code robot descriptions, lack support for arbitrary constraints, and suffer from local minima. Using Drake \cite{drake}, we design excitation trajectories for arbitrary robots and constraints, leveraging a custom black-box augmented Lagrangian solver to mitigate some of the numerical issues that can hinder gradient-based optimization of information-maximizing objectives.

\section{Autonomous Asset Generation Via Robot Interaction}  %

\subsection{Problem Statement}

Our pipeline autonomously generates simulation assets for rigid objects using a standard pick-and-place setup consisting of a robotic manipulator, two bins, and an external RGBD camera. For each object, the pipeline produces a visual geometry $\mathcal{V}$ (a textured visual mesh), collision geometry $\mathcal{C}$ (a union of convex meshes), and physical properties $\mathcal{P}$ (mass, center of mass, rotational inertia). The robot picks an object from the first bin, reconstructs its visual geometry (Section \ref{visual_geometry}), derives a collision geometry (Section \ref{collision_geometry}), identifies physical properties (Section \ref{system_id}), and places the object in the second bin. This process repeats for all objects. Figure \ref{system_diagram} illustrates the asset generation workflow.

\subsection{Geometric Reconstruction for Visual Geometry}
\label{visual_geometry}

We reconstruct the visual geometry $\mathcal{V}$ by collecting multi-view object observations $\{\mathcal{I}\}^N$ (Section \ref{visual_geometry:data_collection}) and using state-of-the-art (SOTA) reconstruction methods to create a textured triangle mesh (Section \ref{visual_geometry:geometric_reconstruction}).

\subsubsection{Data Collection and Processing}
\label{visual_geometry:data_collection}

\begin{figure}
\centerline{
\includegraphics[width=\linewidth, keepaspectratio]
{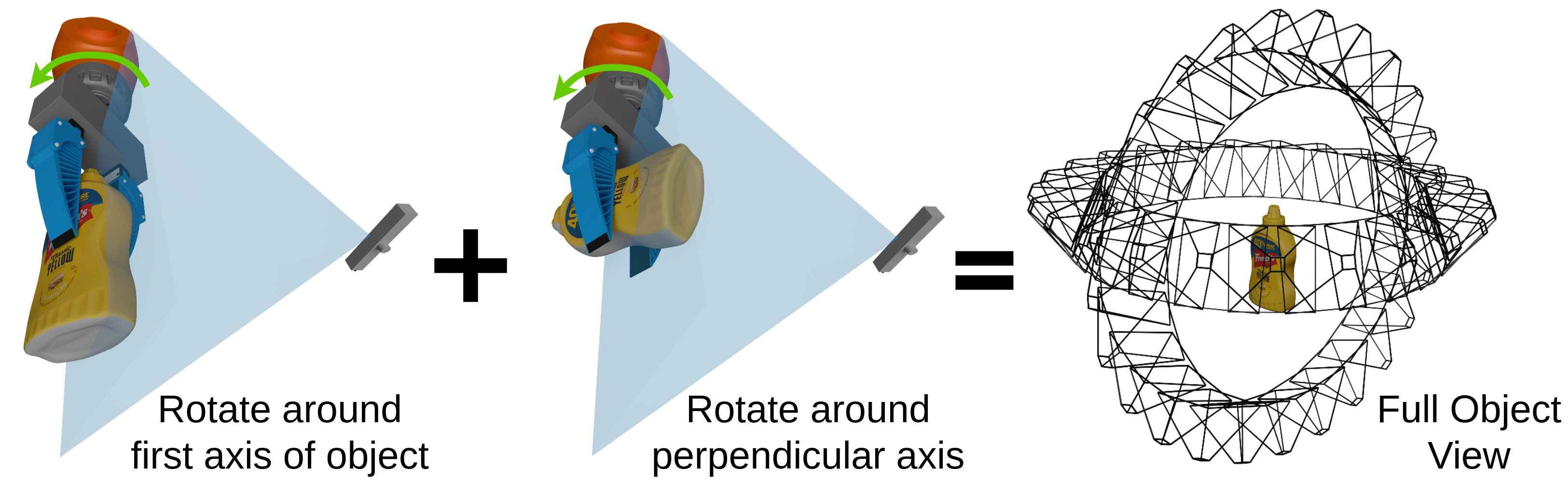}
}
\caption{\textbf{Our object scanning method.} We use re-grasps to display the object along two perpendicular axes, providing the camera with a complete view of the object.}
\label{scanning_diagram}
\end{figure}

The robot rotates the object along two perpendicular axes using primitive motions and re-grasps as needed, ensuring all object surfaces are visible across multiple frames despite occasional gripper occlusions. This provides a full view of the object's surface (Figure \ref{scanning_diagram}). We compute grasp pairs by ranking feasible grasps based on antipodal quality \cite{grasp_pose_detection_in_pcd} and a pair score, which measures how perpendicular and spatially separated two grasps are to avoid mutual occlusions.
The robot collects redundant RGB images ${\mathcal{I}}^{N_L}$ during scanning, which are downsampled to ${\mathcal{I}}^N$ for computational efficiency. Downsampling is performed by selecting every $K^{\text{th}}$ frame, followed by iteratively selecting additional frames from the remaining set. Specifically, we iteratively choose the frame with the largest cosine distance (in DINO \cite{dino} feature space) from the already selected frames until we reach $N$ total images, ensuring maximal diversity in the final selection.
Object and gripper masks $\{\mathcal{M}^O\}^N$ and $\{\mathcal{M}^G\}^N$ are extracted using SAM2 \cite{sam2} video segmentation.

\subsubsection{Geometric Reconstruction}
\label{visual_geometry:geometric_reconstruction}

\begin{figure}
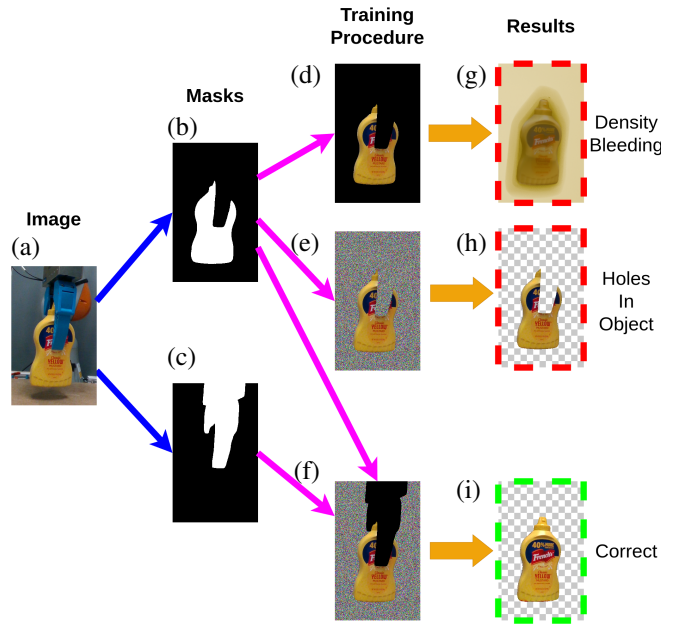

    \centering
    \begin{overpic}[width=\linewidth]{figures/object_centric_training_labels.png}
        \put (0, 57.5) {(a)}
        \put (24, 76) {(b)}
        \put (24, 40.5) {(c)}
        \put (43,84) {(d)}
        \put (43,58) {(e)}
        \put (43.5,23) {(f)}
        \put (68.5,84) {(g)}
        \put (68.5,58) {(h)}
        \put (68.5,20) {(i)}
    \end{overpic}
        \caption{\textbf{Our object-centric visual reconstruction recipe.} From the collected RGB images (a), we obtain the object masks (b) and gripper masks (c). Using only the object masks to ignore background pixels during training (d) results in density bleeding into unoccupied regions (g). Applying alpha-transparent training (e) mitigates density bleeding but incorrectly drives occluded object regions toward transparency (h). Ignoring pixels inside of the gripper mask during training, along with employing alpha transparent training (f), successfully reconstructs an unoccluded object view with no density bleeding (i).}
    \label{object_centric_training}
\end{figure}

We obtain the object's visual mesh and texture map using SOTA implicit reconstruction methods. These methods are typically designed to reconstruct an entire static scene rather than dynamic, object-centric scenarios \cite{pulkit_real2sim}. We propose a general recipe to adapt them for object-centric reconstructions from our dynamic scenes, demonstrating it with three reconstruction approaches. This recipe applies broadly to any method relying on photometric losses.

Standard approaches like NeRF \cite{nerf} assume a static scene with known camera poses. In contrast, our setup features a moving object and a stationary camera whose fixed pose does not need to be known.
To adapt, we redefine the object frame as the world frame, masking out the background so that the non-masked region remains static relative to the new world frame, as in \cite{bundlesdf}. Camera poses $\{\mathcal{X}^C\}^N$ are obtained via object tracking, as the transformation from the camera to the new world frame corresponds to the object's pose in the camera frame.

Using object masks $\{\mathcal{M}^O\}^N$, we train the reconstruction method on pixels belonging to the object. However, excluding background pixels can lead to density bleeding (see Figure \ref{object_centric_training}), where the model assigns nonzero density to empty space due to a lack of supervision. To address this, we employ alpha-transparent training \cite{instantngp}, which enforces zero density in the background without additional hyperparameters.
This method replaces background pixels in the training data with iteration-dependent random colors and blends those same colors into the predicted image based on the model’s density predictions. Since the model cannot predict these random colors, it minimizes the loss by assigning zero density outside the object, allowing the colors to shine through. As part of this work, we integrated alpha-transparent training into Nerfstudio, enabling support for object-centric reconstruction.

Alpha-transparent training resolves background issues but cannot handle occlusions, such as those caused by the gripper, as it would incorrectly make occluded regions transparent. To address this, we use gripper masks $\{\mathcal{M}^G\}^N$ to exclude occluded pixels from the reconstruction objective. Gripper masks take precedence over object masks.

For featureless objects like single-color surfaces, depth supervision, which constrains reconstruction with depth data, improves accuracy by resolving ambiguities in photometric losses. This additional geometric constraint is particularly effective for objects such as bowls, where photometric methods often struggle.

\subsection{Collision Geometries}
\label{collision_geometry}

The visual geometry $\mathcal{V}$ is simplified into a convex collision geometry $\mathcal{C}$ for physics simulation. Following prior works \cite{real2sim2real_hao_su, pulkit_real2sim}, we use approximate convex decomposition algorithms \cite{vhacd, coacd}, which split $\mathcal{V}$ into nearly convex components and simplify each using convex hulls. This process yields a computationally efficient and simulatable geometry $\mathcal{C}$.

We note that simulating a collection of convex pieces can be suboptimal, especially when meshes overlap or contain gaps. In Drake’s hydroelastic contact model \cite{hydroelastic}, gaps can distort the pressure field, causing dynamic artifacts. In the point contact models used in most robotics simulators, overlaps may generate interior contact points.
Another option is to use primitive geometries to represent the collision geometry. For instance, sphere-based approximations might enable rapid simulations on GPU-accelerated simulators.
The optimal representation depends on the simulator and the tradeoff between simulation speed and accuracy. While we found convex decomposition effective in most cases, particularly with point contact models, a more general approach remains an avenue for future work.

\subsection{Physical Parameter Identification}
\label{system_id}

We identify object parameters by first using the robot's joint-torque sensors to determine the robot arm's parameters. Then, we re-identify these parameters with the object grasped and compute the object's parameters as the difference.

\subsubsection{Parameters to Identify}
\label{params_to_identify}

We estimate the object's inertial parameters: mass $m_B$, center of mass $\boldsymbol{p}^{com}_B$, and rotational inertia $\boldsymbol{I}_B \in S_3$ (the set of $3 \times 3$ symmetric matrices). These parameters are physically feasible if and only if the pseudo-inertia $\boldsymbol{J}_B$ of body $B$ is positive definite \cite{physical_consistent_LMI}:
\begin{equation}
    \label{pseudo_inertia}
    \boldsymbol{J}_B := \begin{bmatrix} {\bf \Sigma}_B
      & {\bf h}_B \\ {\bf h}_B^T & m_B \end{bmatrix} \succ 0,
\end{equation}
where $\boldsymbol{h}_B = m_B \cdot \boldsymbol{p}^{com}_B$, $\mathbf{\Sigma}_B = \frac{1}{2}\mathrm{tr}({\bf I}_B)\mathbb{I}_{3 \times 3} - {\bf I}_B$, and $\mathbb{I}_{3 \times 3}$ is the identity matrix.
We leave the estimation of the object's contact parameters as important future work.

For the initial robot identification, we identify the inertial parameters for each link, i.e., $m_{n}$, $\boldsymbol{p}^{com}_{n}$, and $\boldsymbol{I}_{n}$ where $n\in\{1,...,N\}$ is the link index. We also identify the joint friction coefficients $\boldsymbol{\mu}_v \in \mathbb{R}^N \ge \mathbf{0}$, $\boldsymbol{\mu}_c \in \mathbb{R}^N \ge \mathbf{0}$ and the reflected rotor inertia $\boldsymbol{I_r} \in \mathbb{R}^N \ge \mathbf{0}$.

\subsubsection{Robot Identification}
\label{robot_identification}

The robot dynamics follow the manipulator equations \cite{underactuated}:
\begin{equation}
    \label{manipulator_equations}
    \mathbf{M}(\mathbf{q})\mathbf{\ddot{q}} + \mathbf{C}(\mathbf{q},\mathbf{\dot{q}})\mathbf{\dot{q}} - \boldsymbol{\tau_g}(\mathbf{q}) + \boldsymbol{\tau}_f(\mathbf{\dot{q}}) + \boldsymbol{\tau}_r(\mathbf{\ddot{q}}) = \boldsymbol{\tau},
\end{equation}
where the terms (in order) represent the mass matrix, the bias term containing the Coriolis and gyroscopic effects, the torques due to gravity, the torques due to joint friction, and the torques due to reflected rotor inertia.
These equations are affine in the parameters $\boldsymbol{\alpha} \in \mathbb{R}^{13N}$ that comprise the inertial, friction, and reflected inertia parameters for each link.
By measuring joint torques $\boldsymbol{\tau}$ and kinematic states, we solve for $\boldsymbol{\alpha}$ using linear least-squares, forming an overdetermined system:
\begin{equation}
    \boldsymbol{\Tau} = \mathbf{W}\boldsymbol{\alpha} + \mathbf{w}_0.
\end{equation}
This becomes a semi-definite program (SDP) when imposing physical feasibility constraints \cite{physical_consistent_LMI}:
\begin{equation}
    \label{sdp_program}
    \min_{\boldsymbol{\alpha}} ||\mathbf{W}\boldsymbol{\alpha} + \mathbf{w}_0 - \boldsymbol{\Tau}||_2^2 \quad
    s.t. \quad {\bf J}_n \succ 0,
    \boldsymbol{\mu}_v, \boldsymbol{\mu}_c, \boldsymbol{I}_r \ge \mathbf{0},
\end{equation}
where $\boldsymbol{\mu}_v$ and $\boldsymbol{\mu}_c$ are viscous and Coulomb friction coefficients, and $\boldsymbol{I}_r$ represents reflected rotor inertia.
SDPs are convex programs that can be solved to global optimality. In practice, we solve a slightly modified version of this program that excludes unidentifiable parameters \cite{physical_consistent_LMI, geometric_robot_dynamic_id} from the objective and uses the regularization from \cite{geometric_robot_dynamic_id}.

\subsubsection{Optimal Excitation Trajectory Design}
\label{optimal_excitation_trajectory_design}

To identify parameters $\boldsymbol{\alpha}$ effectively, we need high-quality data that sufficiently excites all parameters. Optimal excitation trajectory design aims to maximize the information content in the regressor matrix $\mathbf{W}(\mathbf{q})$ by choosing joint trajectories that provide the most informative data.

For ordinary least-squares (OLS), the parameter estimate $\boldsymbol{\hat{\alpha}}$ minimizes the residual error $\min_{\boldsymbol{\alpha}} ||\mathbf{W}(\mathbf{q})\boldsymbol{\alpha} - \boldsymbol{\tilde{\Tau}}||_2^2$, with the variance of the estimate given by $\text{Var}(\boldsymbol{\hat{\alpha}}) \approx (\mathbf{W}(\mathbf{q})^T\mathbf{W}(\mathbf{q}))^{-1}$. Although our problem involves constrained least-squares, we use the variance properties of OLS as a proxy, aiming to minimize $\text{Var}(\boldsymbol{\hat{\alpha}})$ by maximizing the information content of $\mathbf{W}(\mathbf{q})$.
To balance information across all parameters and reduce worst-case uncertainty, we minimize a weighted combination of the condition number $f_c(\boldsymbol{\lambda}) = \sqrt{\lambda_{\text{max}} / \lambda_{\text{min}}}$, which ensures even information distribution, and the E-optimality criterion $f_e(\boldsymbol{\lambda}) = -\lambda_{\text{min}}$, which minimizes the largest uncertainty \cite{leeOptimalExcitationTrajectories2021}. Here, $\boldsymbol{\lambda}$ are the eigenvalues of $\mathbf{W}(\mathbf{q})^T\mathbf{W}(\mathbf{q})$.
This results in the following optimization problem that searches over data matrices $\mathbf{W}(\mathbf{q})$ while respecting the robot constraints:
\begin{equation}
\label{optimal_experiment_design_program}
\begin{aligned}
\hspace{1.25em}\min_{\mathclap{\mathbf{q}(t), t \in [t_0, t_f]}} & \hspace{1.75em} f_c(\boldsymbol{\lambda}) + \gamma f_e(\boldsymbol{\lambda}), \\
\text{s.t.} & \hspace{1.75em}\text{collision avoidance}, \\
& \hspace{1.75em} \text{joint position/ velocity/ acceleration limits}, \\
& \hspace{1.75em} \text{zero start and end velocities/ accelerations}.
\end{aligned}
\end{equation}
To make the infinite-dimensional problem tractable, we parameterize the trajectories using a finite Fourier series, as is standard practice for excitation trajectory design \cite{birdy, Tian2024ExcitationTO}.

Existing approaches, such as \cite{leeOptimalExcitationTrajectories2021}, rely on gradient-based solvers, which often struggle with local minima when additional constraints are introduced. We suspect that the cost function’s dependence on eigenvalues contributes to numerical issues, making the optimization inherently challenging.
Instead, we use the augmented Lagrangian method, which reformulates constrained optimization as a series of unconstrained subproblems with penalty terms enforcing constraints. The underlying numerical challenges of the objective make these subproblems difficult to solve, and hard constraints like collision avoidance further exacerbate the issue. To address this, we develop a custom augmented Lagrangian solver inspired by \cite{lancelot} but solve the subproblems using black-box optimization.
This approach improves solution quality at the expense of longer runtimes, an acceptable trade-off for an offline process solved once per environment.
Our Drake implementation supports arbitrary robots using standard robot description formats such as URDFs and allows for arbitrary constraints through its \textit{MathematicalProgram} interface. This addresses a key practical limitation of the more narrow implementations in existing opens-source alternatives \cite{birdy, leeOptimalExcitationTrajectories2021, Tian2024ExcitationTO}.

\subsubsection{Object Identification}
\label{object_id}
After a one-time identification of the robot's parameters, we re-identify the last link's parameters with the object grasped. The object's parameters are computed as the difference $p$ between the two parameter sets, leveraging the lumped parameter linearity of composite bodies:
\begin{equation}
    m_{12} = m_1 + m_2,
    \quad \boldsymbol{h}_{12} = \boldsymbol{h}_1 + \boldsymbol{h}_2,
    \quad \boldsymbol{I}_{12} = \boldsymbol{I}_1 + \boldsymbol{I}_2.
\end{equation}
The physically feasible set of lumped parameters is closed under addition but not under subtraction. In theory, subtracting one reasonable parameter set from another, such as subtracting a sub-body's parameters from those of a composite body, should yield a valid parameter set. However, estimation errors can lead to physically invalid results, such as a small object's mass falling within the error margin, resulting in a negative mass estimate. Hence, we add a pseudo-inertia constraint ${\bf J}_p \succ 0$ for the parameter difference $p$, which we found to be informative in practice.
We identify the object using the same excitation trajectory as for robot identification, as designing a trajectory specific to the last link proved less effective, and using the same trajectory mitigates systematic errors when subtracting parameters.
We assume a rigid grasp and leave slippage for future work.

\section{Implementation Details}
\label{implementation_details}

In this section, we go over some of the key implementation details. The complete codebase is available at \url{https://github.com/nepfaff/scalable-real2sim}.

\begin{figure}
\centerline{
\includegraphics[width=\linewidth, keepaspectratio]
{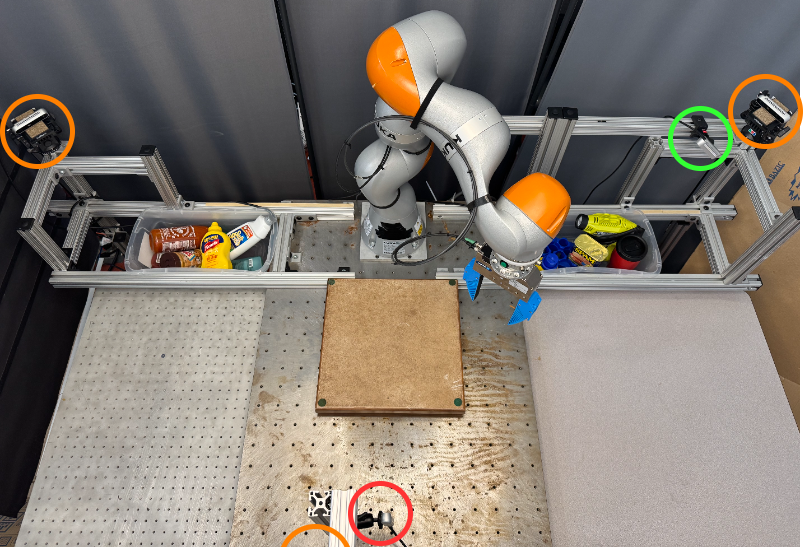}
}
\caption{\textbf{Our pick-and-place setup.} It features a Kuka LBR iiwa 7 arm with a Schunk WSG-50 gripper and Toyota Research Institute Finray fingers. Workspace observations rely on three static RealSense D415 cameras (orange circles), while bin picking uses a RealSense D435 (green circle), and object scanning is performed with another D415 (red circle). All cameras capture 640×480 resolution RGBD images. Objects are picked from the right bin and placed into the left bin. A platform is used in the scanning workspace to enhance the iiwa’s kinematic range during re-grasping.}
\label{real_robot_setup}
\end{figure}

\subsubsection{Grasp and Motion Planning}

We implement antipodal grasping based on \cite{grasp_pose_detection_in_pcd} and use \cite{werner2024superfast} for motion planning.

\subsubsection{Object Tracking}

We use BundleSDF \cite{bundlesdf} with a static camera for object tracking, leveraging the filtered frames $\{\mathcal{I}\}^N$ and object masks $\{\mathcal{M}^O\}^N$ to compute camera poses $\{\mathcal{X}^C\}^N$. BundleSDF produces a visual geometry as a byproduct of tracking.

\subsubsection{Geometric Reconstruction}

We applied our recipe from Section \ref{visual_geometry:geometric_reconstruction} to three methods: Nerfacto from Nerfstudio \cite{nerfstudio}, Gaussian Frosting \cite{guedon2024frosting}, and Neuralangelo \cite{neuralangelo}.

\subsubsection{Physical Parameter Identification}
\label{implementaiton_details_system_id}

We collect 10 trajectories for robot identification, averaging positions, and torques to improve the signal-to-noise ratio. Object identification uses a single trajectory for practicality. Velocities and accelerations are computed via double differentiation, and all quantities are low-pass filtered, with cutoff frequencies tuned via hyperparameter search to minimize Equation \ref{sdp_program}.

To handle variations in gripper pose, we identify the robot with multiple gripper openings and match the closest one during object identification. Without this correction, gripper pose variations would introduce errors in the system identification process.
We use iterative closest point (ICP) \cite{icp} to align the point cloud and the known grasp position with the reconstructed mesh, allowing us to express the inertial parameters in the object frame.

\section{Results}
\label{results}

\subsection{Geometric Reconstruction}

\begin{figure*}
\centerline{
\includegraphics[width=\linewidth, keepaspectratio]
{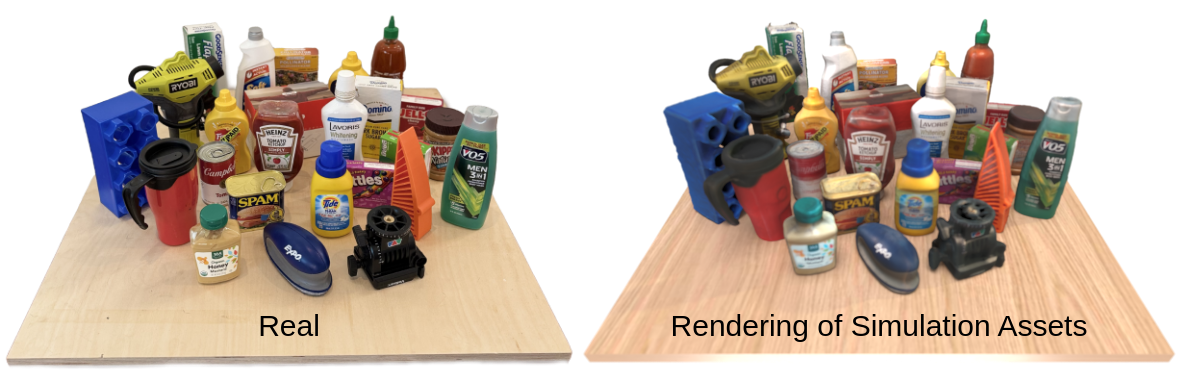}
}
\caption{\textbf{Real-world objects (left) and their reconstructed counterparts (right).} Each object on the left was individually reconstructed using our pipeline. These assets were then manually arranged in simulation to approximately match their real-world poses and rendered to produce the image on the right. The strong visual similarity is notable, especially given that the reconstructions are rendered triangle meshes rather than neural renders.}
\label{real_object_figure}
\end{figure*}

\begin{figure}
\centerline{
\includegraphics[width=1\linewidth, keepaspectratio]
{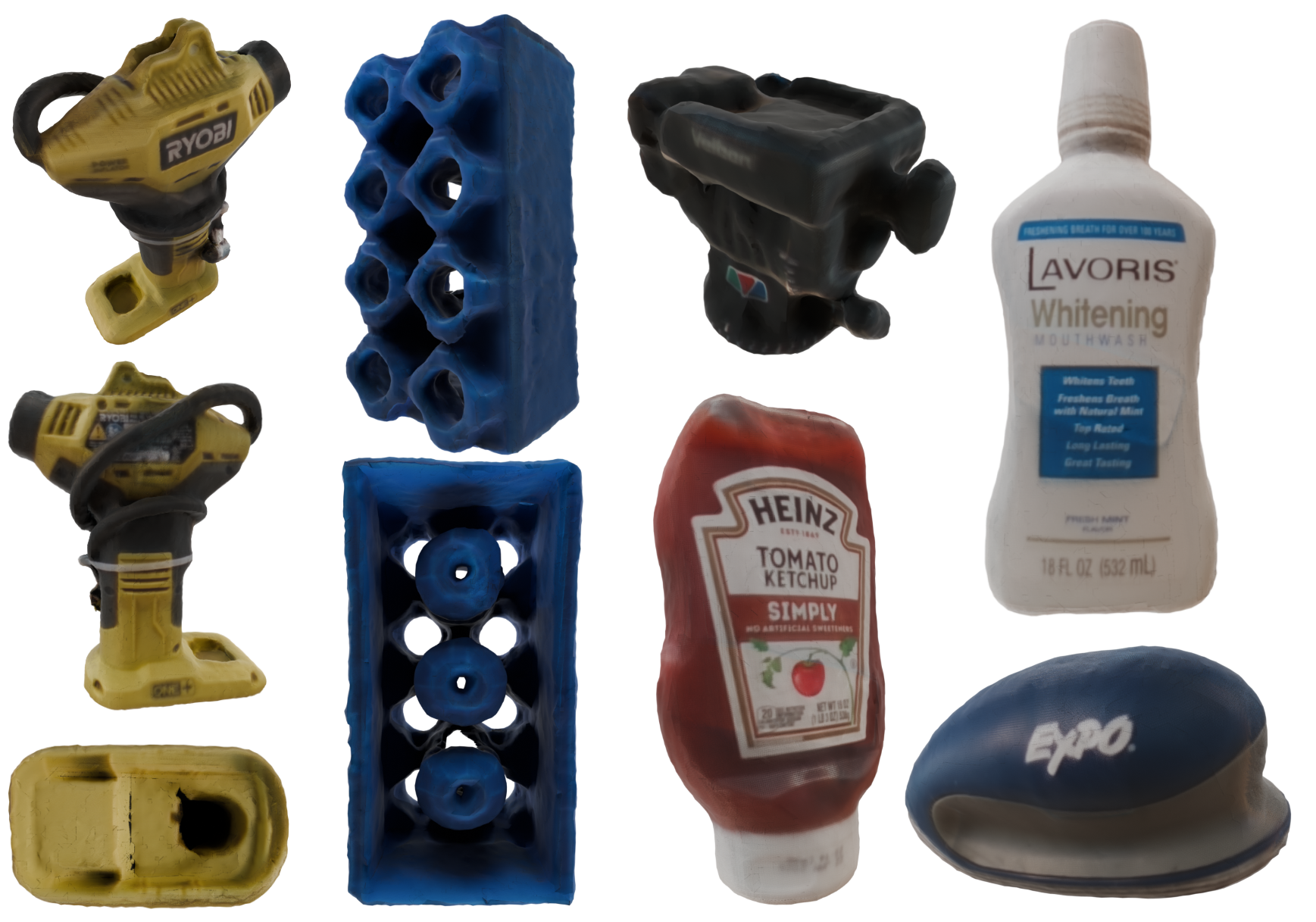}
}
\caption{\textbf{A selection of geometric reconstructions.} The first two columns show multiple views of the same object (a power inflator in the first column and a Lego block in the second), demonstrating the completeness of our reconstructions. The last two columns highlight close-up views of other objects, illustrating the accuracy of both geometric and visual reconstruction, even for parts that were occluded during scanning. We provide interactive 3D visualizations on our \href{https://scalable-real2sim.github.io/}{project page}.}
\label{geometric_reconstructions}
\end{figure}

\begin{figure}
\centerline{
\includegraphics[width=1\linewidth, keepaspectratio]
{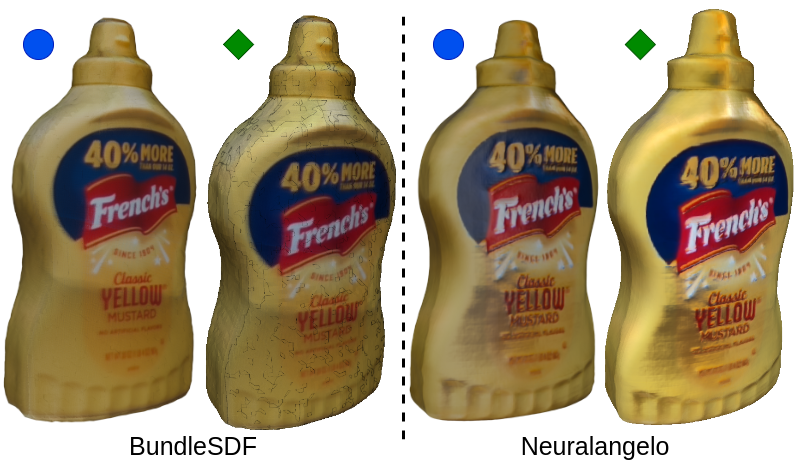}
}
\caption{\textbf{Comparison of BundleSDF \cite{bundlesdf} and Neuralangelo \cite{neuralangelo} reconstructions of a mustard bottle.} Blue circles denote Blender \cite{blender} renders, while green diamonds represent Meshlab \cite{meshlab} renders. The BundleSDF mesh appears best in Blender but worst in Meshlab due to poor topology (e.g., scattered boundary faces), which requires a powerful renderer to compensate. In contrast, the Neuralangelo mesh maintains consistent quality across both renderers due to its well-structured topology. The effects of poor topology in the BundleSDF mesh appear as black lines, which originate from the mesh itself rather than the texture map. These artifacts are particularly noticeable at the top of the bottle's body.}
\label{bundlesdf_vs_neuralangelo}
\end{figure}

We assess reconstruction accuracy by computing the Chamfer distance between our BundleSDF reconstructions of selected YCB \cite{ycb} objects and their corresponding 3D scanner models from the original dataset. Our method achieves reconstruction errors of 0.93mm, 1.68mm, 5.58mm, and 0.80mm for the mustard bottle, potted meat can, bleach cleanser, and gelatin box, respectively. These low errors indicate that our system produces both accurate and complete object reconstructions. However, if the physical product dimensions have changed since the dataset’s release, the measured reconstruction error may be artificially inflated.
Qualitative results are provided in Figures \ref{real_object_figure} and \ref{geometric_reconstructions}. Notably, Figure \ref{geometric_reconstructions} highlights our ability to reconstruct entire objects, including previously occluded regions such as the bottom, which would be inaccessible without object interaction.
We found that BundleSDF produces meshes with poor topology, including scattered boundary faces and non-manifold vertices. These artifacts complicate rendering and may pose challenges for simulators that require watertight meshes. One way to mitigate rendering issues is to use a high-quality but slow renderer like Blender Cycles \cite{blender}. Alternatively, a higher-quality mesh can be generated using a SOTA reconstruction method like Neuralangelo, following our geometric reconstruction recipe from Section \ref{visual_geometry:geometric_reconstruction}. Figure \ref{bundlesdf_vs_neuralangelo} presents a qualitative comparison between BundleSDF and Neuralangelo, highlighting the rendering artifacts caused by poor topology.
Additional examples, including Nerfacto and Gaussian Frosting reconstructions, are available on our \href{https://scalable-real2sim.github.io/}{project page}.

\subsection{Physical Parameter Identification}

We replicate the test object and benchmark from \cite{fast_object_inertial_id}. Their test object enables the creation of objects with varying inertial properties of known ground truth values. Unlike \cite{fast_object_inertial_id}, we do not use a force-torque sensor and thus directly mount the test object to the robot's last link.

\begin{table}[h]
\centering
\caption{Comparison of identification errors between our method and results reported in \cite{fast_object_inertial_id}. `FT` indicates that the results were obtained with a force-torque sensor, while `JT` refers to joint-torque sensors. The metrics are the ones proposed in \cite{fast_object_inertial_id}. Error bars represent one standard deviation.}
\label{object_id_metric_table}
\begin{tabular}{lccc}
\toprule
 & Mass (\%) & CoM (\%) & Inertia Tensor (\%) \\
\midrule
OLS (FT) & 3.18 & 24.9 & $>$500 \\
PMD-25K (FT) & 4.45 & 4.50 & \textbf{29.11} \\
Ours (JT) & \textbf{1.34} $\pm$ 0.21 & \textbf{2.15} $\pm 1.20 $ & 42.35 $\pm$ 15.64 \\
\bottomrule
\end{tabular}
\end{table}

Following the identification procedure from \ref{system_id}, we identify the same eight test object configurations as \cite{fast_object_inertial_id}, performing three identifications per configuration and averaging the metrics over all 24 trials.
Table \ref{object_id_metric_table} shows our results compared to the best-performing method (`PMD-25K`) from \cite{fast_object_inertial_id} and their OLS baseline. Notably, our method does not require external sensors or prior knowledge of object shape and pose, unlike `PMD-25K`. While our approach achieves superior accuracy in mass and center of mass estimation, it performs slightly worse for rotational inertia. However, it still significantly outperforms OLS across all metrics.
Additionally, estimating inertial parameters from joint torque sensors is more challenging than using a force-torque sensor, as they only provide indirect force measurements.

To validate our approach further, we conduct an end-to-end experiment by 3D printing a mustard bottle with known inertia and processing it through the pipeline. The ground truth mass is 427.1g. Our estimation errors are 1.23\% for mass, 6.33\% for the center of mass, and 358.6\% for rotational inertia, based on the metrics from \cite{fast_object_inertial_id}. While the rotational inertia error is notably high, these results remain encouraging, given the challenges introduced by our camera calibration errors and limits from joint torque sensing.

\subsection{Simulation Performance}

\begin{figure}
\centerline{
\includegraphics[width=1\linewidth, keepaspectratio]
{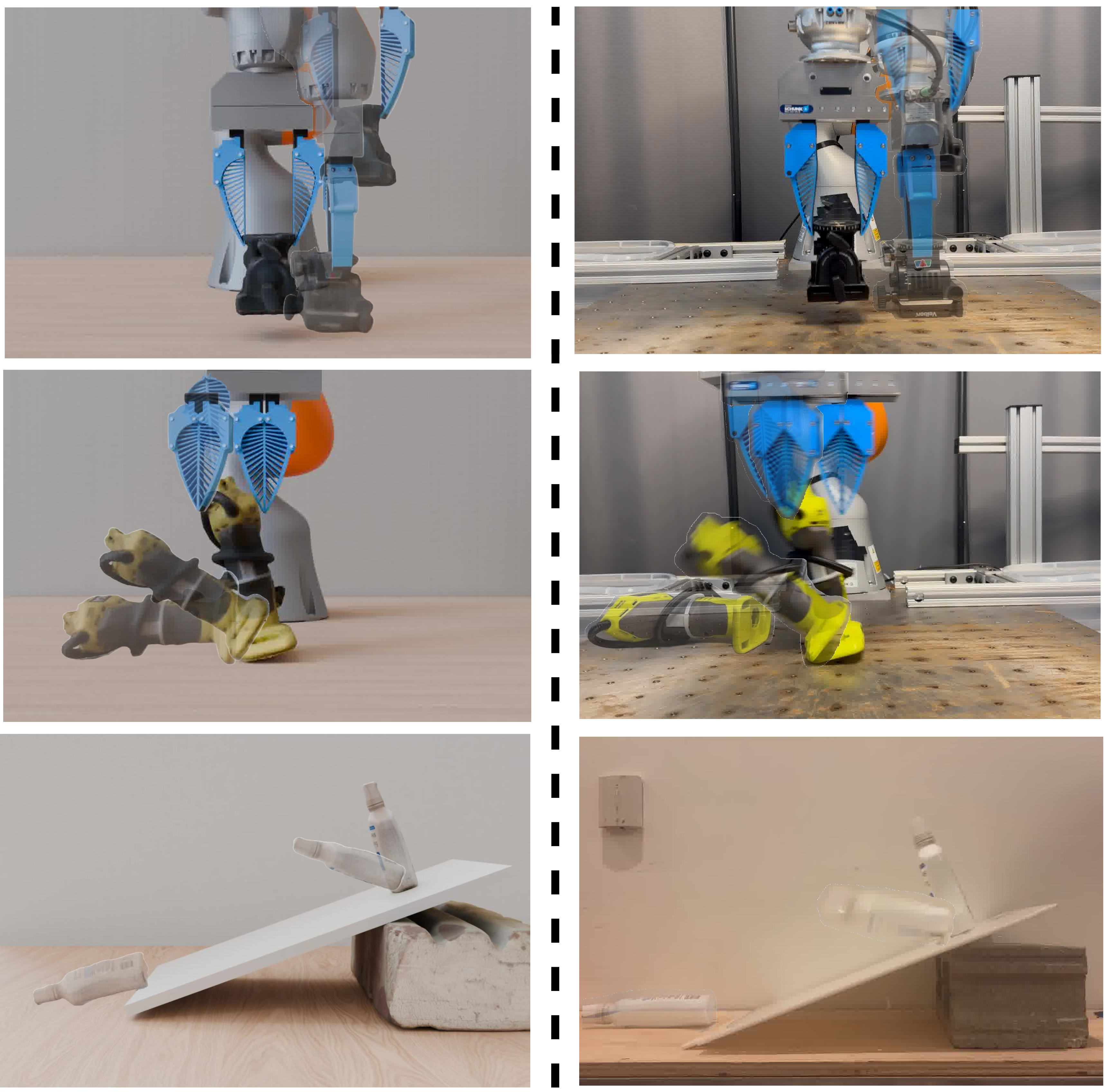}
}
\caption{\textbf{Our simulation experiments.} The left column represents simulations, and the right column represents their real-world counterparts. The first row is pick-and-place, the second is knocking over, and the third is falling down a ramp. Different frames are overlaid transparently to show motion, and videos are available on the \href{https://scalable-real2sim.github.io/}{project page}.}
\label{simulation_experiments}
\end{figure}

Our pipeline produces simulatable assets that can be directly imported into physics simulators such as Drake. However, individual simulation rollouts are highly sensitive to initial conditions, making direct one-to-one comparisons with real-world rollouts challenging. A robust evaluation would require comparing the distributions of rollout trajectories rather than individual instances.
One approach to mitigate sensitivity to initial conditions is to compute equation error metrics \cite[Chapter~18]{underactuated} by resetting the simulation state to match the real-world state at every timestep. However, this would require a precise state estimation system, which is beyond the scope of this work. Instead, we focus on qualitative evaluation, presenting interactive simulations and side-by-side comparisons of real-world and simulated rollouts on our \href{https://scalable-real2sim.github.io/}{project page}. See Figure \ref{simulation_experiments} for an overview.

\addtolength{\textheight}{-1cm}   %

\section{Conclusion}

This paper introduced an automated Real2Sim pipeline that generates simulation-ready assets for real-world objects through robotic interaction. By automating the creation of object geometries and physical parameters, our approach eliminates the need for manual asset generation, addressing key bottlenecks in Sim2Real. We demonstrated that our method accurately estimates object geometry and physical properties without manual intervention, enabling scalable dataset creation for simulation-driven robotics research.

\section*{Acknowledgement}

This work was supported by Amazon.com, PO No. 2D-15693043, and the Office of Naval Research (ONR) No. N000142412603.
We thank Lirui Wang, Andy Lambert, Thomas Cohn, Ge Yang, and Ishaan Chandratreya for their discussions.

\bibliographystyle{IEEEtran}
\bibliography{ref}

\end{document}